\documentclass[letterpaper]{article} % DO NOT CHANGE THIS
\usepackage{aaai25}  % DO NOT CHANGE THIS
\usepackage{times}  % DO NOT CHANGE THIS
\usepackage{helvet}  % DO NOT CHANGE THIS
\usepackage{courier}  % DO NOT CHANGE THIS
\usepackage[hyphens]{url}  % DO NOT CHANGE THIS
\usepackage{graphicx} % DO NOT CHANGE THIS
\usepackage{natbib}  % DO NOT CHANGE THIS AND DO NOT ADD ANY OPTIONS TO IT
\usepackage{caption} % DO NOT CHANGE THIS AND DO NOT ADD ANY OPTIONS TO IT
\usepackage{algorithm}
\usepackage{algorithmic}

\usepackage{multirow}
\usepackage{amsmath}
\usepackage{bbding}
\usepackage{xcolor}
\definecolor{customBlue}{HTML}{2952FF}
\definecolor{customPurple}{HTML}{991456}
\usepackage{colortbl} 
\usepackage{amsfonts}
\usepackage{booktabs}

\usepackage{newfloat}
\usepackage{listings}
\DeclareCaptionStyle{ruled}{labelfont=normalfont,labelsep=colon,strut=off} % DO NOT CHANGE THIS
\floatstyle{ruled}
\newfloat{listing}{tb}{lst}{}
\floatname{listing}{Listing}
\def\logo{\makebox[0pt][l]{\hspace{-6pt}\raisebox{-1ex}{\includegraphics[height=30pt]{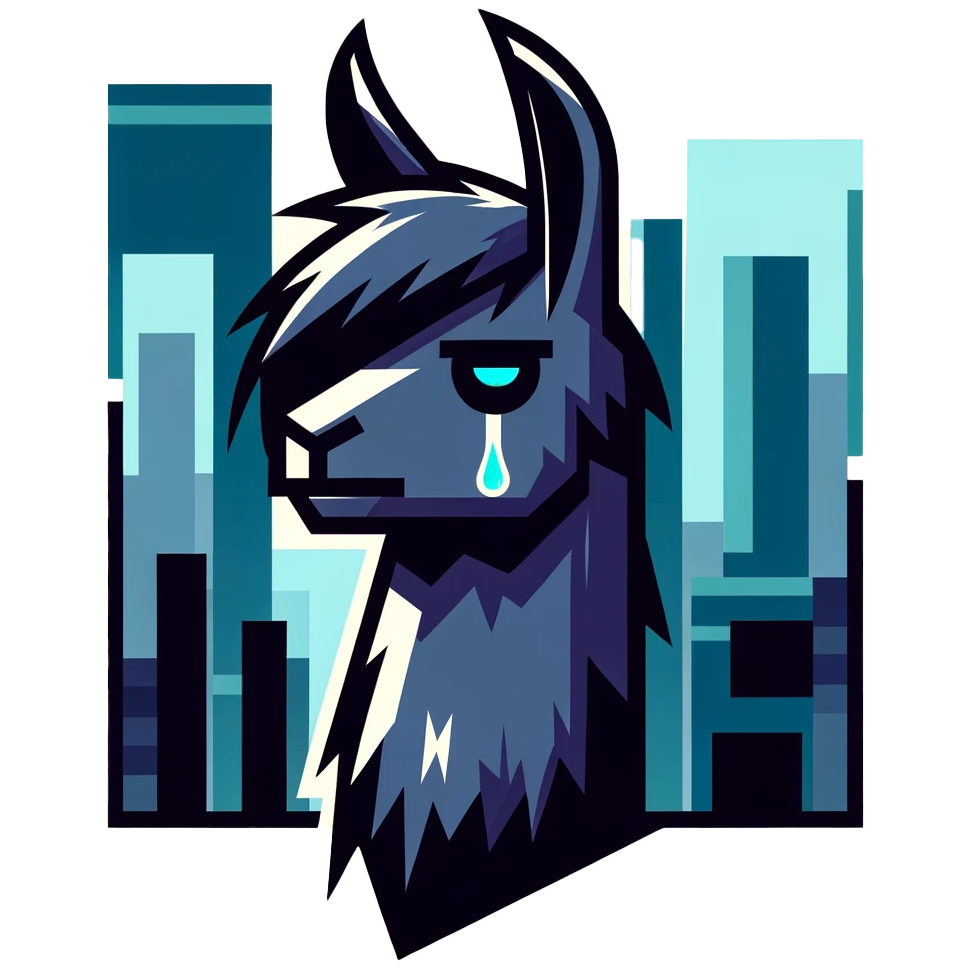}}}}
\title{\vspace{-1.2ex}\logo \ \ \ \ \ \ EMO-LLaMA: Enhancing Facial Emotion Understanding with Instruction Tuning}
\author{
    Bohao Xing\equalcontrib\textsuperscript{\rm 1}, 
    Zitong Yu\equalcontrib\thanks{Corresponding authors: Xin Liu and Zitong Yu.}\textsuperscript{\rm 2},
    Xin Liu\footnotemark[2]\textsuperscript{\rm 1},
    Kaishen Yuan\textsuperscript{\rm 3},
    Qilang Ye\textsuperscript{\rm 2},
    Weicheng Xie\textsuperscript{\rm 4},
    Huanjing Yue\textsuperscript{\rm 3},
    Jingyu Yang\textsuperscript{\rm 3},
    Heikki Kälviäinen\textsuperscript{\rm 1}
}

\affiliations{
    \textsuperscript{\rm 1}Lappeenranta-Lahti University of Technology LUT \quad
    \textsuperscript{\rm 2}Great Bay University \quad
    \textsuperscript{\rm 3}Tianjin University \quad
    \textsuperscript{\rm 4}Shenzhen University
}

\usepackage{bibentry}

\begin{document}

\maketitle

\begin{abstract}
Facial expression recognition (FER) is an important research topic in emotional artificial intelligence. In recent decades, researchers have made remarkable progress. 
However, current FER paradigms face challenges in generalization, lack semantic information aligned with natural language, and struggle to process both images and videos within a unified framework, making their application in multimodal emotion understanding and human-computer interaction difficult. 
Multimodal Large Language Models (MLLMs) have recently achieved success, offering advantages in addressing these issues and potentially overcoming the limitations of current FER paradigms.
However, directly applying pre-trained MLLMs to FER still faces several challenges. Our zero-shot evaluations of existing open-source MLLMs on FER indicate a significant performance gap compared to GPT-4V and current supervised state-of-the-art (SOTA) methods. 
In this paper, we aim to enhance MLLMs’ capabilities in understanding facial expressions. 
We first generate instruction data for five FER datasets with Gemini. We then propose a novel MLLM, named EMO-LLaMA, which incorporates facial priors from a pretrained facial analysis network to enhance human facial information. 
Specifically, we design a Face Info Mining module to extract both global and local facial information. Additionally, we utilize a handcrafted prompt to introduce age-gender-race attributes, considering the emotional differences across different human groups. 
Extensive experiments show that EMO-LLaMA achieves SOTA-comparable or competitive results across both static and dynamic FER datasets.
The instruction dataset and code are available at \url{https://github.com/xxtars/EMO-LLaMA}.
\end{abstract}

\vspace{-1em}
\section{Introduction}
Facial expressions reflect a person’s emotional state and are vital for effective interpersonal interactions. Understanding the emotional states from facial expressions is increasingly significant due to its applications, such as human-computer interaction~\cite{liu2017facial}, healthcare aids~\cite{bisogni2022impact}, and driving safety~\cite{wilhelm2019towards}.

\begin{figure}
    \centering    
    \includegraphics[width=0.8\linewidth]{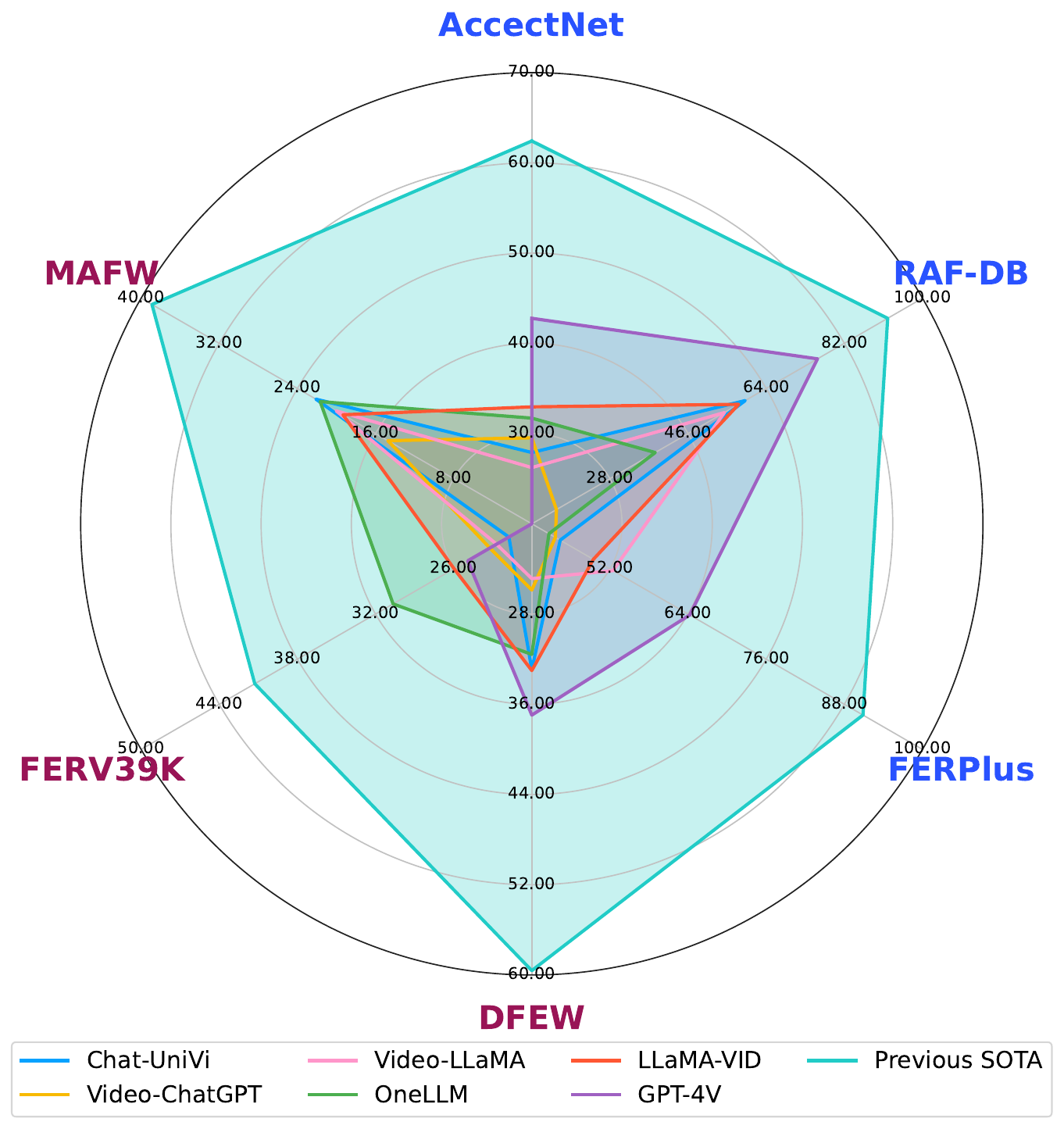}
    \vspace{-0.1em}
    \caption{Comparison of zero-shot and previous supervised SOTA on several FER datasets. \textcolor{customBlue}{\textbf{Blue}} represents image dataset, and \textcolor{customPurple}{\textbf{Purple}} represents video dataset.}
    \vspace{-2em}
    \label{fig:zeroshot}
\end{figure}

In recent decades, researchers have significantly advanced by expanding datasets and developing more efficient architectures. However, existing paradigms for FER face the following challenges: 
i) Existing methods struggle with generalization across different datasets and modalities~\cite{suresh2021critically, chen2021cross}.
ii) FER is typically divided into static facial expression recognition (image) and dynamic facial expression recognition (video), and current paradigms struggle to handle both tasks in a unified framework.
iii) Existing paradigms focus on close-set recognition and lack semantic understanding and often overlook other semantic cues. Applying these paradigms to multimodal emotion understanding and human-computer interaction remains a significant challenge.

MLLMs have gained significant popularity in the natural language processing and computer vision fields, offering advantages in generalization and natural language interaction. These advantages of MLLMs hold promise for overcoming the limitations of current FER paradigms. 

However, directly applying pre-trained MLLMs to FER still faces several challenges. As shown in Fig.~\ref{fig:zeroshot} or the table provided in the \textbf{Appendix} (due to space constraints), extensive zero-shot evaluations of existing open-source MLLMs on FER were conducted, revealing that current open-source MLLMs struggle with emotion understanding and significantly lag behind the most advanced closed-source MLLM, GPT-4V~\cite{achiam2023gpt, lian2023gpt}. They have a considerable gap compared to the current supervised SOTA methods.

Additionally, some work has attempted to leverage their rich prior knowledge for text-based emotion understanding~\cite{lei2023instructerc, zhang2023dialoguellm, liu2024emollms}. Despite their success, the application of MLLMs for emotion understanding in images or videos remains under-explored. Given that facial expressions are crucial emotional cues in human face-to-face communication, this work aims to enhance MLLMs’ ability to understand facial expressions and establish a basis for future multimodal, multi-cued emotion understanding tasks. 

Specifically, we aim to enhance the emotion understanding capabilities of existing MLLMs through improvements in the FER task, thereby narrowing the gap between MLLM-based approaches and traditional classification paradigms. However, there exist three major challenges in the FER task when deploying MLLMs: 
i) There are no suitable FER instruction datasets. Existing FER datasets have either coarse-grained FER labels~\cite{li2017reliable, barsoum2016training, mollahosseini2017affectnet, jiang2020dfew, wang2022ferv39k} or limited emotion descriptions~\cite{liu2022mafw}, and directly using these to construct instruction data would limit the diversity of LLM responses. 
ii) Current FER paradigms utilize pre-cropped face images or video frames, which could diminish the MLLMs’ general understanding abilities and their sensitivity to other visual cues. This also limits the potential for extending MLLMs to multimodal emotion understanding in the future.
iii) Furthermore, visual features from vision encoders like CLIP~\cite{radford2021learning} of current MLLMs struggle to capture facial information, and leaving the impact of facial priors on enhancing MLLMs for the FER task unexplored. We plan to utilize potentially useful information, such as facial embedding, landmarks~\cite{tautkute2018know}, and age-gender-race attributes which considers that different races and age groups express emotions in distinct ways~\cite{smith2002race, shields2013gender, wilkins2014class, dailey2010evidence}.

To address these challenges, we propose an instruction-tuning approach for the FER task. We select commonly used static and dynamic FER datasets and generated suitable instruction tuning datasets with Gemini~\cite{team2023gemini}. This instruction dataset enhances the diversity of manually constructed recognition instructions using only emotion labels. Additionally, we introduce a novel MLLM, \textit{EMO-LLaMA}, specifically designed for the FER task by incorporating facial priors into a pre-trained MLLM like LLaMA-VID~\cite{li2023llama}. To obtain facial prior knowledge, we utilize a pre-trained face encoder and decoder to extract three types of facial features: facial embedding, landmarks, and age-gender-race attributes, which complement the general vision encoder. This work provides new insights for both the emotion and MLLM research communities. 
Our main contributions are summarized as follows:

\begin{itemize}
    \vspace{-0.2em}
    \item To the best of our knowledge, this is the first attempt to unify image and video FER tasks using instruction tuning over MLLMs, which is challenging for traditional paradigms.
    \vspace{-0.2em}
    \item We utilize Gemini to generate an instruction dataset for five publicly available FER datasets. Specifically, it includes 295k data for image modality and 45k data for the video modality, which is a substantial amount of instructional data. This will provide significant benefits to both the MLLM and FER communities.
    \vspace{-0.2em}
    \item To efficiently train on FER tasks and leverage facial prior knowledge, we introduce the EMO-LLaMA model, which involves tuning LoRA on a pre-trained MLLM and incorporating three types of facial priors. Specifically, we design a Face Info Mining module to extract both global and local facial information. Additionally, we utilize a handcrafted prompt to introduce age-gender-race attributes, considering the emotional differences across different human groups.
    \vspace{-0.2em}
    \item We demonstrate the effectiveness of EMO-LLaMA on six FER datasets, showing that EMO-LLaMA achieves SOTA-comparable or competitive performance. In addition, our experiments demonstrate the generalization capabilities of our approach, which are lacking in current paradigms.
\end{itemize}

\vspace{-1em}
\section{Related work}

\textbf{Facial Expression Recognition.} Currently, FER can be roughly divided into two types: Static Facial Expression Recognition (SFER) and Dynamic Facial Expression Recognition (DFER). SFER~\cite{wen2023distract, li2022emotion, savchenko2022classifying, chen2023static} mainly focuses on recognizing expressions from static images, whereas DFER~\cite{sun2023mae, sun2024hicmae, chen2023static, sun2023svfap} concentrates on recognizing expressions from dynamic image sequences or videos. Traditional methods typically handle SFER and DFER independently, lacking the ability to address facial expression recognition within a unified framework. Additionally, most existing FER frameworks are primarily focused on classification tasks and face challenges in generalization~\cite{suresh2021critically, chen2021cross} and lack semantic information aligned with natural language~\cite{li2023cliper, zhao2023prompting,yuan2024auformer}. Additionally, some research suggests that different races and cultures exhibit variations in the facial expressions of emotions, but existing methods have not taken this into account~\cite{dailey2010evidence, benitez2018multicultural, matsumoto1992american}. These factors limit their applicability to emotion reasoning or multimodal emotion understanding tasks. 

\begin{figure*}
  \centering
  \includegraphics[width=\linewidth]{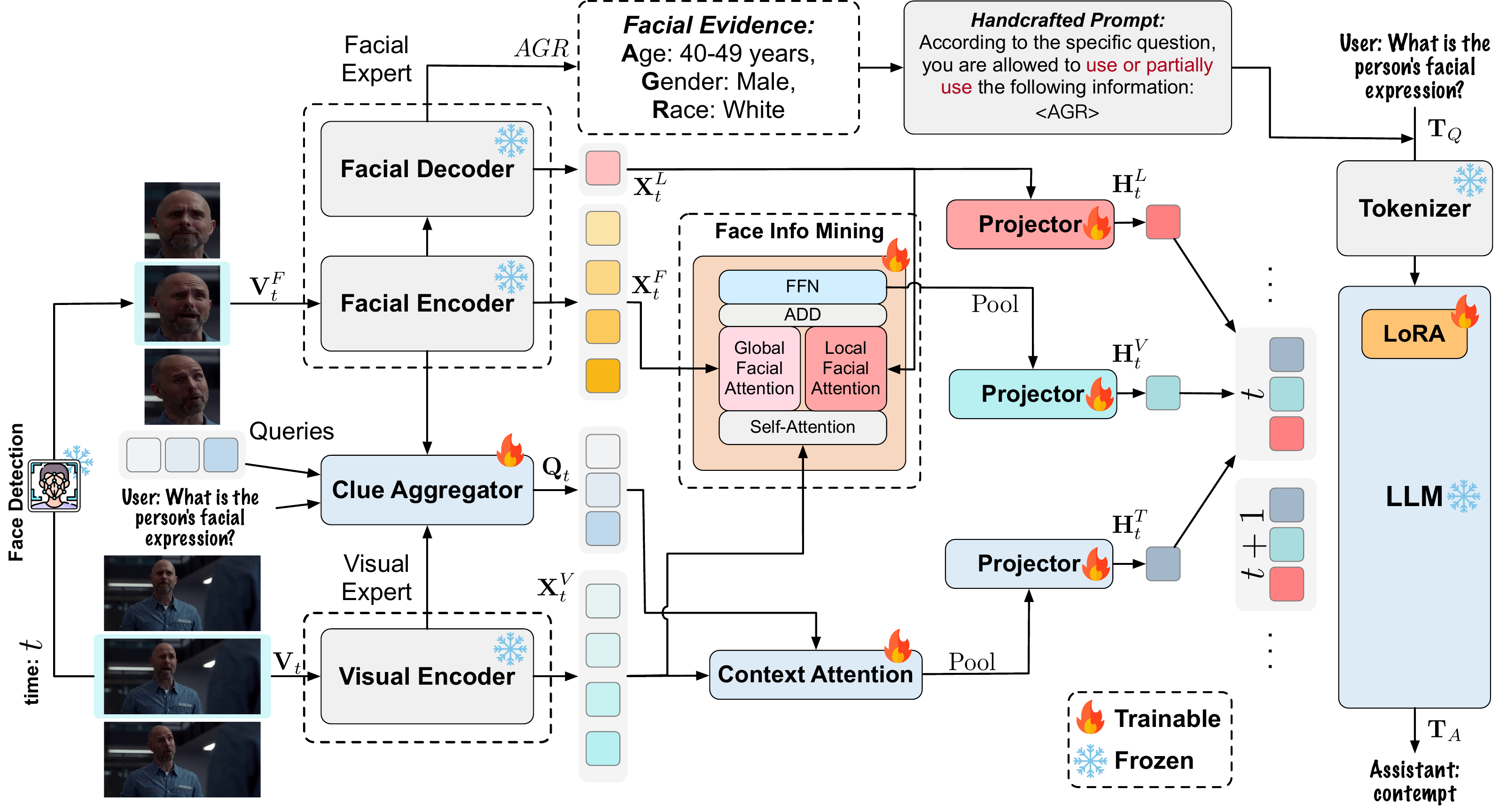}
  \vspace{-1em}
  \caption{The framework of \raisebox{-2pt}{\includegraphics[height=10pt]{Figure/EMOMLLM.png}} EMO-LLaMA. The model first obtains a face image from a face detection network, which is then fed into a facial analysis expert to extract facial prior knowledge. A Clue Aggregator extracts task-specific embedding. General visual features are enhanced by facial features in the Face Info Mining module. Additionally, landmark embedding and handcrafted prompts of facial evidence are further utilized to enhance face information as input to the LLM.}
  \label{fig:framework}
  \vspace{-1.5em}
\end{figure*}

\textbf{Multimodal Large Language Models.} MLLMs are getting popular in multi-modal content understanding, e.g., images~\cite{liu2024visual, wang2023cogvlm, liu2023improved, chen2023lion, chen2024far, zhu2023minigpt} and videos~\cite{zhang2023video, jin2023chat, ataallah2024minigpt4, lin2023video, maaz2023video, tang2024avicuna, li2023llama, he2024ma, song2023moviechat}. These models are built on top of LLMs~\cite{chiang2023vicuna, jiang2023mistral, touvron2023llama, bai2023qwen} and transform visual (videos and images) and textual data into sequences of tokens for input, resulting in generative modeling of downstream multimodal understanding tasks through next-token prediction. Specifically, for image-based MLLMs, image tokens are typically encoded using CLIP~\cite{radford2021learning, sun2023eva}. Similarly, video tokens are encoded using CLIP with or without temporal modeling modules or by utilizing dedicated video encoders~\cite{girdhar2023imagebind, zhu2023languagebind}.

\textbf{Emotion Understanding with LLMs.} There have been initial explorations into text-based emotion understanding using LLMs. DialogueLLM~\cite{zhang2023dialoguellm} leverages GPT-4V~\cite{achiam2023gpt} to extract visual information and generate textual descriptions, then combines them with contextual content to address the task of emotion recognition in conversations. EmoLLMs~\cite{liu2024emollms} focused on both classification and regression tasks in sentiment analysis and introduced a multi-task affective analysis instruction dataset for instruction tuning. More relevant to us are AffectGPT~\cite{lian2023explainable} and EmoLA~\cite{li2024facial}. Although AffectGPT~\cite{lian2023explainable} attempted to address the emotion reasoning task, it lacks sufficient data due to the difficulty of annotation, having conducted only an initial exploration with 100 data samples. Additionally, AffectGPT did not incorporate model designs related to tasks. EmoLA~\cite{li2024facial} relates to facial action units~\cite{ekman1978facial, zhi2020comprehensive} and SFER, but it did not align with the metrics of traditional paradigms and was tested only on the RAF-DB~\cite{li2017reliable}, lacking extensive validation across a broad array of image and video FER datasets. 
One potential approach, similar to HuggingGPT~\cite{shen2024hugginggpt}, is to directly integrate existing paradigms with LMMs. Although this has not yet been explored, we believe it would still face the inherent limitations of current paradigms.

\vspace{-1em}
\section{EMO-LLaMA}

In this section, we present EMO-LLaMA, a multimodal large language model with facial priors. Fig.~\ref{fig:framework} illustrates an overview of the EMO-LLaMA framework. Beyond the general-purpose MLLM with a vision encoder, projector, and large language model, our approach integrates facial priors to enhance the MLLM’s capability of extracting facial information.
Specifically, a frame $\textbf{V}_t \in \mathbb{R}^{H \times W \times 3}$ at time $t$ in a video (or an image) is encoded by a visual encoder known as CLIP~\cite{radford2021learning} to produce the visual embedding $\textbf{X}^V_t \in \mathbb{R}^{N \times C}$. Here, $N = \frac{H}{p} \times \frac{W}{p}$ and $C$ indicate the number of image patches and embedding channels, respectively. The patch size $p$ is typically set to 14 for ViT-based backbones~\cite{dosovitskiy2020image, radford2021learning}. Notably, the visual embedding $\textbf{X}^V_t$ may fail to capture facial structure information because CLIP is trained with general image-text pairs rather than facial-related datasets.

\subsection{Progressive Incorporation of Facial Prior Knowledge}

To incorporate facial prior knowledge into MLLMs, we make use of a pre-trained facial expert to extract facial priors and progressively integrate them into the existing MLLM.

\textbf{Facial Priors Extraction.}
To obtain a face image $\textbf{V}^F_t \in \mathbb{R}^{H^F \times W^F \times 3}$ suitable for processing by a facial expert, we employ a pre-trained face detector, MTCNN~\cite{zhang2016joint}~\footnote{\url{https://github.com/timesler/facenet-pytorch}}, to detect and crop the face region \textit{online}. 
Then, we adopt FaceXFormer~\cite{narayan2024facexformer}\footnote{\url{https://github.com/Kartik-3004/facexformer}} for facial analysis, which includes face parsing, landmark detection, head pose estimation, attribute recognition, and the estimation of age, gender, race, and landmark visibility.
In this work, we utilize facial embedding from the encoder, landmarks, as well as age, gender, and race attributes from the decoder, integrating them with the existing MLLM. Other facial priors can also be considered and will be explored in future work.
Specifically, the facial embedding feature extracted by the facial encoder $f^{ENC}_{face}(\cdot)$ can be expressed as: 
\begin{equation}
\textbf{X}^F_t = f^{ENC}_{face}(\textbf{V}^F_t).
\end{equation}Since FaceXFormer is trained on tasks like face parsing and landmark detection, the facial embedding $\textbf{X}^F_t \in \mathbb{R} ^{N^F \times C}$ includes more structural and low-level details, which CLIP lacks. 

\textbf{Clue Aggregator.} 
We employ a Q-Former~\cite{dai2024instructblip} to dynamically capture instruction-aware visual and facial hidden features to enrich fine-grained clues. 
Specifically, Q-Former takes the user instruction $\textbf{T}_Q$, learnable queries $\textbf{Q} \in \mathbb{R} ^{M \times C}$, general visual embedding $\textbf{X}^V_t$ and facial embedding $\textbf{X}^F_t$ as input, where $M$ denotes the number of queries, typically set to 32. During training, the module enhances task-specific feature extraction as query $\textbf{Q}_t$. To reduce the burden of visual tokens on the LLM, we follow LLaMA-VID~\cite{li2023llama} to generate $\hat{\textbf{H}}^T_t \in \mathbb{R} ^{1 \times C}$ by a cross attention and pool operation as: 
\begin{equation}
\hat{\textbf{H}}^T_t = \operatorname{Pool}(\operatorname{Softmax}(\textbf{Q}_t {\textbf{X}^V_t}^\top)\textbf{X}^V_t),
\end{equation}where the $\operatorname{Softmax}$ and $\operatorname{Pool}$ are conducted along the $N$ and $M$ dimensions, respectively. 

\begin{figure}
    \centering
    \includegraphics[width=\linewidth]{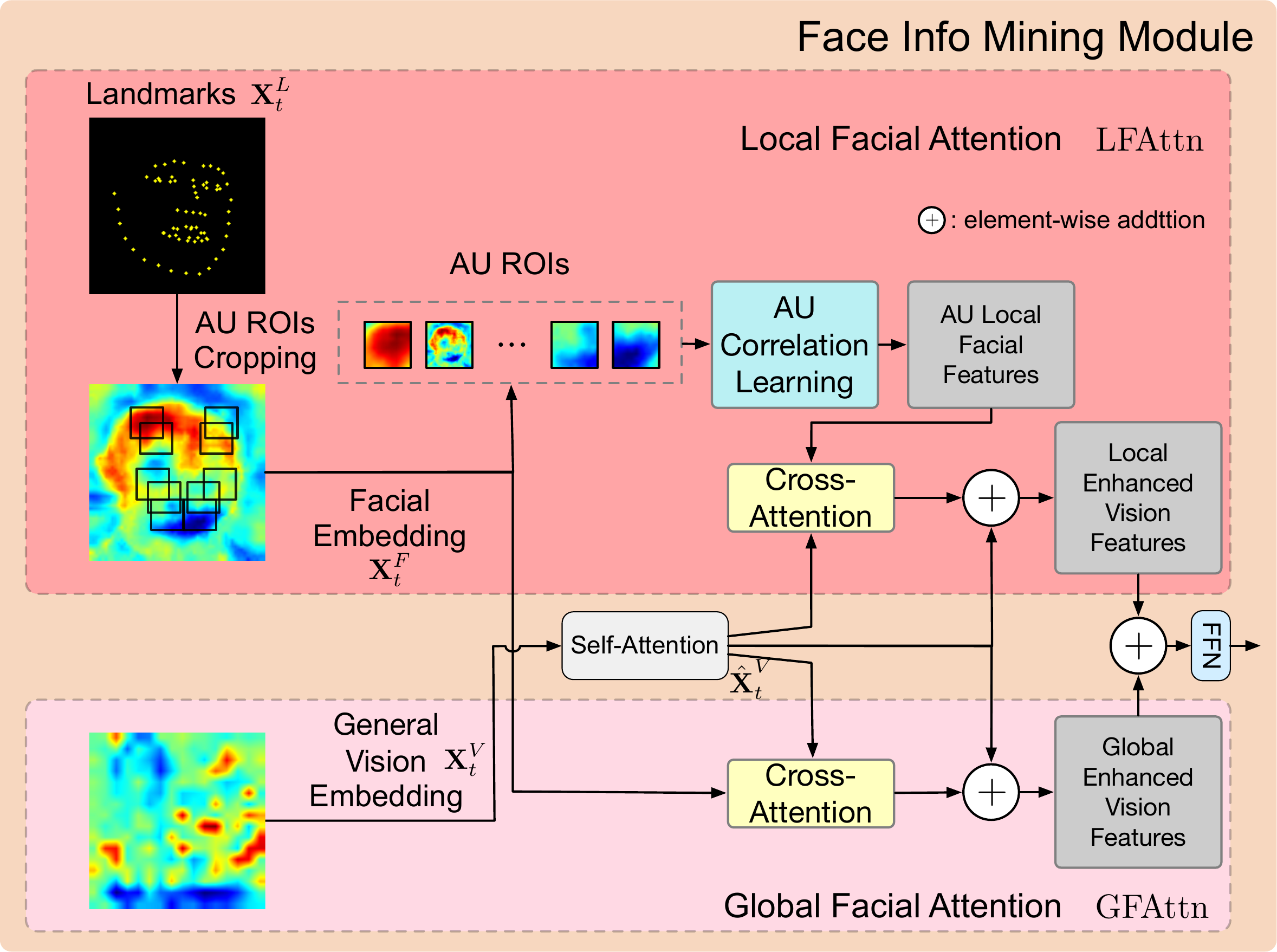}
    \vspace{-1.5em}
    \caption{Face Info Mining Module.}
    \vspace{-1.5em}
    \label{fig:au_local}
\end{figure}

\textbf{Face Info Mining.} 
Additionally, we introduce a Face Info Mining module to enhance the general vision embedding with global and local facial information, as shown in Fig.~\ref{fig:au_local}. 
To maintain the efficiency of LLMs by limiting the number of final visual tokens and to effectively capture the structural information from facial embedding, we plan to interact general embedding and facial embedding before pooling. In particular, the face info mining process can be formulated as:
% \begin{equation}
% \hat{\textbf{H}}^V_t = \operatorname{Pool}(\operatorname{FFN}(\operatorname{XAttn}(\operatorname{SAttn}(\textbf{X}^V_t), \textbf{X}^F_t))),
% \end{equation}
\begin{equation}
\begin{aligned}
\hat{\textbf{H}}^V_t = \operatorname{Pool}(&\operatorname{FFN}(\operatorname{GFAttn}(\operatorname{SAttn}(\textbf{X}^V_t), \textbf{X}^F_t)) \\
& +\operatorname{LFAttn}(\operatorname{SAttn}(\textbf{X}^V_t), \textbf{X}^F_t, \textbf{X}^L_t))), 
\end{aligned}
\end{equation}where the $\operatorname{SAttn}$ and $\operatorname{FFN}$ are self attention and feed-forward network~\cite{vaswani2017attention}. $\operatorname{GFAttn}$ and $\operatorname{LFAttn}$ are designed Global Face Attention and Local Face Attention, which are used to extract global and local facial information. The $\operatorname{LFAttn}$ leverages Action Units~\cite{liu2023multi} to extract local information. More equations and details about the Face Info Mining module can be found in the \textbf{Appendix}.

\textbf{Landmark Embedding and Handcrafted Prompt of Facial Evidence.} 
Furthermore, we utilize the facial prior information extracted by the decoder $f^{DEC}_{face}(\cdot)$, specifically focusing on landmarks $\textbf{X}^L_t$ and \textbf{A}ge-\textbf{G}ender-\textbf{R}ace $AGR$, expressed as:
\begin{equation}
\textbf{X}^L_t, AGR = f^{DEC}_{face}(f^{ENC}_{face}(\textbf{V}^F_t)).
\end{equation}Instead of directly incorporating facial evidence into the instruction, we design a handcrafted prompting method to guide the model in selectively using the $AGR$ information. This approach helps avoid potential negative influences from imperfect predictions by FaceXFormer. 
We use the prompt: ``\textit{According to the specific question, you are allowed to use or partially use the following information:} $AGR$'' to instruct the LLM whether or not to use the $AGR$ information to answer the current question.

\begin{table*}
 \vspace{-0.5em}
 \caption{Performance comparison on facial expression recognition with previous SOTA. \XSolidBrush represents modality mismatch, and - represents unreported results.}
 \vspace{-1em}
  \label{tab:sota}
  \centering
  \small
  \begin{tabular*}{\textwidth}{@{\extracolsep{\fill}}l@{\extracolsep{\fill}}c@{\extracolsep{\fill}}c@{\extracolsep{\fill}}c@{\extracolsep{\fill}}c@{\extracolsep{\fill}}c@{\extracolsep{\fill}}c@{\extracolsep{\fill}}c@{\extracolsep{\fill}}c@{\extracolsep{\fill}}c@{\extracolsep{\fill}}}
    \toprule
    \multirow{2}{*}{Methods} & \multicolumn{1}{c}{AffectNet} & \multicolumn{1}{c}{RAF-DB} & \multicolumn{1}{c}{FERPlus} & \multicolumn{2}{c}{DFEW} & \multicolumn{2}{c}{FERV39K} & \multicolumn{2}{c}{MAFW} \\ 
    \cmidrule(rl){2-2} \cmidrule(rl){3-3} \cmidrule(rl){4-4} \cmidrule(rl){5-6} \cmidrule(rl){7-8} \cmidrule(rl){9-10}
    & Acc & Acc & Acc & UAR & WAR & UAR & WAR & UAR & WAR \\
    \midrule
    SCN~\cite{wang2020suppressing} & 60.23 & 88.14 & 89.35 & \XSolidBrush & \XSolidBrush & \XSolidBrush & \XSolidBrush & \XSolidBrush & \XSolidBrush \\
    EAC ~\cite{zhang2022learn} & - & 88.99 & 89.64 & \XSolidBrush & \XSolidBrush & \XSolidBrush & \XSolidBrush & \XSolidBrush & \XSolidBrush \\
    MVT~\cite{li2021mvt} & 61.40 & 88.62 & 89.22 & \XSolidBrush & \XSolidBrush & \XSolidBrush & \XSolidBrush & \XSolidBrush & \XSolidBrush \\
    APViT~\cite{xue2022vision} & - & \underline{91.98} & \underline{90.86} & \XSolidBrush & \XSolidBrush & \XSolidBrush & \XSolidBrush & \XSolidBrush & \XSolidBrush \\
    PF-ViT~\cite{li2022emotion} & \underline{62.42} & 91.07 & 90.18 & \XSolidBrush & \XSolidBrush & \XSolidBrush & \XSolidBrush & \XSolidBrush & \XSolidBrush \\
    \midrule
    I3D~\cite{carreira2017quo} & \XSolidBrush & \XSolidBrush & \XSolidBrush & 46.52 & 58.27 & 30.17 & 38.78 & - & - \\
    Former-DFER~\cite{zhao2021former} & \XSolidBrush & \XSolidBrush & \XSolidBrush & 53.69 & 65.70 & 37.20 & 46.85 & - & - \\
    EST~\cite{liu2023expression} & \XSolidBrush & \XSolidBrush & \XSolidBrush & 53.43 & 65.85 & - & - & - & - \\
    IAL~\cite{li2023intensity} & \XSolidBrush & \XSolidBrush & \XSolidBrush & 55.71 & 69.24 & 35.82 & 48.54 & - & - \\
    CLIPER~\cite{li2023cliper} & 61.98 & 91.61 & - & 57.56 & 70.84 & 41.23 & 51.34 & - & - \\
    DFER-CLIP~\cite{zhao2023prompting} & \XSolidBrush & \XSolidBrush & \XSolidBrush & \underline{59.61} & \underline{71.25} & \underline{41.27} & \underline{51.65} & \underline{38.89} & \underline{52.55} \\
    \midrule
    EMO-LLaMA (Ours) & \textbf{63.01} & 86.41 & 81.35 & \textbf{60.23} & 65.89 & 38.92 & 47.76 & \textbf{41.57} & 48.63 \\
    \bottomrule
  \end{tabular*}
  \vspace{-2em}
\end{table*}

\begin{figure}
  \centering
  \includegraphics[width=\linewidth]{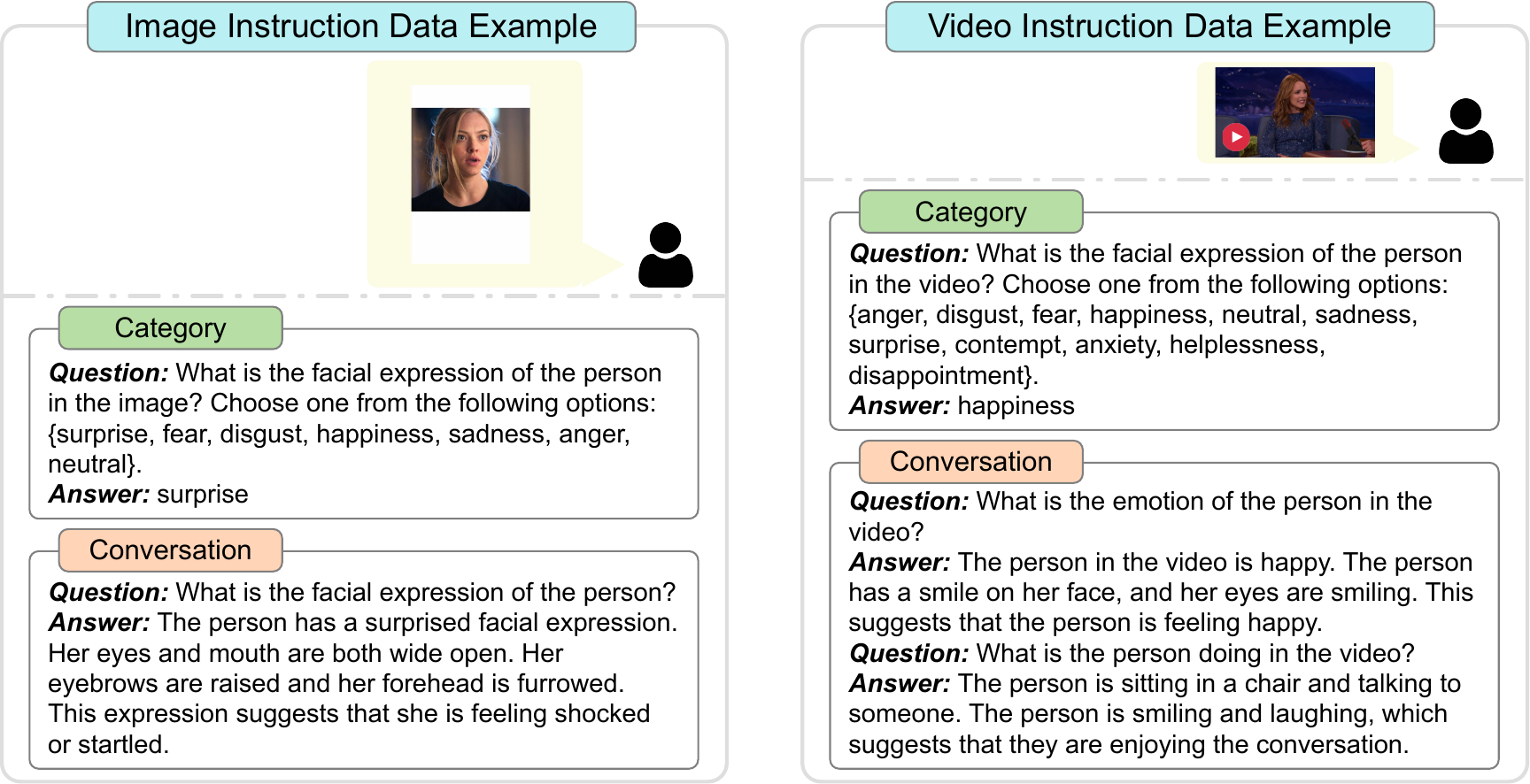}
  \vspace{-1.5em}
  \caption{The examples of our generated instruction data. Zoom in for a closer look, and more examples can be found in the \textbf{Appendix}.}
  \label{fig:instruction_example}
  \vspace{-1.5em}
\end{figure}

\textbf{Instruction Tuning with LoRA}.
After obtaining pooled Q-Former embedding $\hat{\textbf{H}}^T_t$, 
pooled visual embedding $\hat{\textbf{H}}^V_t$, and landmark embedding $\textbf{X}^L_t$, we utilize the multi-layer perceptron (MLP) projector, to project these features to the token embedding space:
\begin{equation}
\textbf{H}^T_t = \operatorname{MLP}^T(\hat{\textbf{H}}^T_t),\, \textbf{H}^V_t = \operatorname{MLP}^V(\hat{\textbf{H}}^V_t),\, \textbf{H}^L_t = \operatorname{MLP}^L({\textbf{X}}^L_t).
\end{equation}

After obtaining the Q-Former embedding token $\textbf{H}^T_t$, the visual embedding token $\textbf{H}^V_t$, and landmark prior token $\textbf{H}^F_t$, we concatenate them together to represent the frame at time $t$. Along with frames at other timestamps, the entire video sequence is translated into the language space in token format, which is then used to generate responses from LLMs.

In this paper, our work is built upon LLaMA-VID~\cite{li2023llama}, although other MLLMs could also be used.
Considering that our instruction dataset is task-specific, we utilize LoRA~\cite{hu2021lora} for fine-tuning. The overall parameters to be optimized are $\boldsymbol{\Theta}$, as shown in Fig.~\ref{fig:framework} with \includegraphics[height=0.8em]{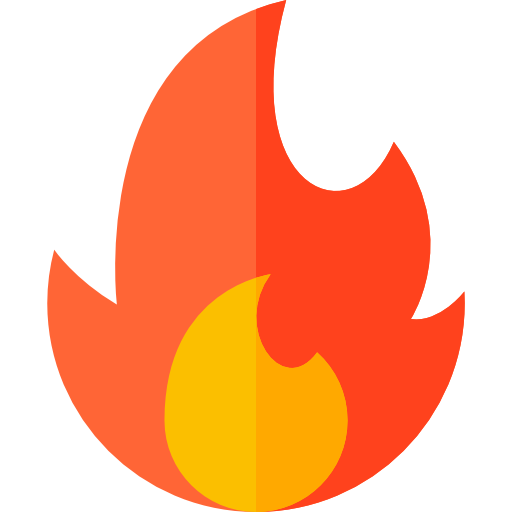}. We optimize these parameters following the autoregressive way, and the likelihood of the target output $\textbf{T}_A$ conditioned on video $\textbf{V}$, facial priors and questions $\textbf{T}_Q$ is given by:
\begin{equation}
\begin{aligned}
p(\textbf{T}_A \mid & \textbf{V}, \textbf{X}^F, \textbf{X}^L, AGR, \textbf{T}_Q) = \\
& \prod_{i=1}^{L} p_{\boldsymbol{\Theta}}(\boldsymbol{x}_i \mid \textbf{V}, \textbf{X}^F, \textbf{X}^L, AGR, \textbf{T}_{Q, <i}, \textbf{T}_{A, <i}),
\end{aligned}
\end{equation}where $\textbf{T}_{Q, <i}$ and $\textbf{T}_{A, <i}$ are the instruction and answer tokens in all turns before the current prediction token $\boldsymbol{x}_i$, respectively.

\subsection{Gemini-assisted Emotion Visual Instruction Data Generation}

We first collect currently publicly available FER datasets, including FERPlus~\cite{barsoum2016training}, RAF-DB~\cite{li2017reliable}, AffectNet~\cite{mollahosseini2017affectnet}, DFEW~\cite{jiang2020dfew}, FERV39K~\cite{wang2022ferv39k}, and MAFW~\cite{liu2022mafw}. 
Then, we use the \textit{Gemini 1.0 Pro Vision} API~\cite{team2023gemini} to generate conversation instructions, further enriching the instruction dataset to enhance the MLLM’s response to face-related questions. Since the topic of FER Visual Instruction Tuning is still in its infancy, no guidelines have been proposed yet for constructing emotion instruction data. We follow the approach of LLaVA~\cite{liu2024visual}, leveraging its recent successes with machine-generated instructions.

\begin{table*}
  \vspace{-0.5em}
  \caption{Evaluation of EMO-LLaMA on the video modality trained with image modality data. \textit{general} means pre-training on general purpose images or videos, \textit{video} refers to pre-training on FER-related videos, and \textit{image} refers to pre-training on FER-related images. More zero-shot results can be found in the \textbf{Appendix}.}
  \vspace{-1em}
  \label{tab:crossmodality_video}
  \centering
  \small
  \begin{tabular*}{0.9\textwidth}{@{\extracolsep{\fill}}l@{\extracolsep{\fill}}c@{\extracolsep{\fill}}c@{\extracolsep{\fill}}c@{\extracolsep{\fill}}c@{\extracolsep{\fill}}c@{\extracolsep{\fill}}c@{\extracolsep{\fill}}c@{\extracolsep{\fill}}}
    \toprule
    \multirow{2}{*}{Methods} & \multirow{2}{*}{Training Modality} & \multicolumn{2}{c}{DFEW} & \multicolumn{2}{c}{FERV39K} & \multicolumn{2}{c}{MAFW} \\ \cmidrule(rl){3-4} \cmidrule(rl){5-6} \cmidrule(rl){7-8} 

    & & UAR & WAR & UAR & WAR & UAR & WAR \\ 
    \midrule
    \multicolumn{8}{c}{\textit{Zero-shot of Contrastive Learning Methods}} \\
    CLIP~\cite{radford2021learning, foteinopoulou2023emoclip} & \textit{general} & 19.86 & 10.60 & 20.99 & 17.10 & 20.04 & 21.00 \\
    EmoCLIP~\cite{foteinopoulou2023emoclip} & \textit{video} & 36.76 & 46.27 & 26.73 & 35.30 & 25.86 & 33.49 \\
    \midrule
    \multicolumn{8}{c}{\textit{Zero-shot of MLLMs}} \\
    Video-LLaMA~\cite{zhang2023video} & \textit{general} & 24.86 & 35.03 & 17.81 & 22.71 & 20.02 & 31.00 \\
    Chat-UniVi~\cite{jin2023chat} & \textit{general} & 32.93 & 37.29 & 21.76 & 25.68 & 22.11 & 27.19 \\
    LLaMA-VID~\cite{foteinopoulou2023emoclip} & \textit{general} & 33.01 & 38.36 & 25.97 & 30.55 & 19.33 & 23.11 \\
    GPT-4V~\cite{achiam2023gpt,lian2023gpt} & - & 36.96 & 43.80 & 24.92 & 34.19 & - & - \\
    \midrule
    EMO-LLaMA (Ours) & \textit{image} & \textbf{44.73} & \textbf{50.45} & \textbf{32.42} & \textbf{37.69} & \textbf{27.26} & \textbf{38.09} \\
    \bottomrule
  \end{tabular*}
  \vspace{-1em}
\end{table*}
\begin{table*}[t]
\centering
\begin{minipage}[t]{0.49\textwidth}
  \caption{Evaluation of EMO-LLaMA on the image modality trained with video modality data.}
  \vspace{-1em}
  \label{tab:crossmodality_image}
  \centering
  \small
  \begin{tabular*}{\textwidth}{@{\extracolsep{\fill}}l@{\extracolsep{\fill}}c@{\extracolsep{\fill}}c@{\extracolsep{\fill}}c@{\extracolsep{\fill}}c@{\extracolsep{\fill}}}
    \toprule
    \multirow{2}{*}{Methods} & Training & AffectNet & RAF-DB & FERPlus \\ \cmidrule(rl){3-3} \cmidrule(rl){4-4} \cmidrule(rl){5-5} 
    & Modality & Acc & Acc & Acc \\
    \midrule
    \multicolumn{5}{c}{\textit{Zero-shot of MLLMs}} \\
    % LLaMA-VID~\cite{li2023llama} & \textit{general} & 32.96 & 57.66 & 49.45 \\
    % GPT-4V~\cite{achiam2023gpt,lian2023gpt} & - & \textbf{42.77} & \textbf{75.81} & \textbf{64.25} \\
    Video-LLaMA & \textit{general} & 26.23 & 54.37 & 52.46 \\
    Chat-UniVi & \textit{general} & 27.90 & 59.16 & 44.36 \\
    LLaMA-VID & \textit{general} & 32.96 & 57.66 & 49.45 \\
    GPT-4V & - & \textbf{42.77} & \textbf{75.81} & \textbf{64.25} \\
    \midrule
    EMO-LLaMA & \textit{video} & 38.43 & 71.51 & 59.29 \\
    \bottomrule
  \end{tabular*}
\end{minipage}
\hfill % Adds horizontal space between the minipages
\begin{minipage}[t]{0.49\textwidth}
  \caption{Zero-shot evaluation of EMO-LLaMA on the extra emotion datasets.}
  \vspace{-1em}
  \label{tab:crossmodality_other}
  \centering
  \small
  \begin{tabular*}{\textwidth}{@{\extracolsep{\fill}}l@{\extracolsep{\fill}}c@{\extracolsep{\fill}}c@{\extracolsep{\fill}}c@{\extracolsep{\fill}}c@{\extracolsep{\fill}}}
    \toprule
    \multirow{2}{*}{Methods} & MELD & MER & CMMA & CASME \uppercase\expandafter{\romannumeral2} \\ \cmidrule(rl){2-2} \cmidrule(rl){3-3} \cmidrule(rl){4-4} \cmidrule(rl){5-5} 
    & UAR & UAR & UAR & Acc \\
    \midrule
    \multicolumn{5}{c}{\textit{Zero-shot of MLLMs}} \\
    % LLaMA-VID~\cite{li2023llama} & 16.28 & 37.19 & 16.82 & 12.89 \\
    % GPT-4V~\cite{achiam2023gpt,lian2023gpt} & - & - & - & 14.64 \\
    Video-LLaMA & 12.64 & 25.85 & 5.12 & 10.36 \\
    Chat-UniVi &  9.57 & 31.85 & 6.00 & 11.27 \\
    LLaMA-VID & 16.28 & 37.19 & 16.82 & 12.89 \\
    GPT-4V & - & - & - & 14.64 \\
    \midrule
    EMO-LLaMA & \textbf{20.39} & \textbf{54.42} & \textbf{22.76} & \textbf{25.58} \\
    \bottomrule
  \end{tabular*}
\end{minipage}
\vspace{-1.5em}
\end{table*}

We use different pipelines for images and videos. For images, we utilize Gemini to generate objects and facial-related questions, uniquely providing facial expression labels and the image. When designing FER-related conversations, Gemini is prompted to base them on the provided facial expression labels. For videos, we additionally provide the central frame and a description generated per second by LLaVA to Gemini. Similarly, when designing FER-related conversations, they are based on the provided facial expression labels. Due to issues such as API calls, the final number of instructions does not exactly match the number of original data. Notably, we do not generate instructions for FERPlus because it consists of grayscale images with a resolution of 48, which we believe have limited utility.
More details about the pipeline visualization, as well as statistics of the collected data, prompts for LLaVA and Gemini, and generated instruction data, can be found in the \textbf{Appendix}. Finally, we generated 295k data for the image modality and 45k for the video modality, providing a substantial amount of instructional data. In Fig.~\ref{fig:instruction_example}, we present two examples of the generated instruction data. 

\vspace{-0.5em}
\section{Experiments}

\begin{table*}[t]
\vspace{-0.5em}
\centering
\begin{minipage}[t]{0.49\textwidth}
      \caption{Ablation study on facial priors.}
      \vspace{-1em}
      \label{tab:facial_prior}
    \centering
      \small
      \begin{tabular*}{\textwidth}{@{\extracolsep{\fill}}c@{\extracolsep{\fill}}c@{\extracolsep{\fill}}c@{\extracolsep{\fill}}c@{\extracolsep{\fill}}c@{\extracolsep{\fill}}}
        \toprule
        \multirow{2}{*}{Facial Embedding} & \multirow{2}{*}{Landmark} & \multirow{2}{*}{ARG} & \multicolumn{2}{c}{MAFW} \\ \cmidrule(rl){4-5}
        & & & UAR & WAR  \\ 
        \midrule
        & & & 34.75 & 44.86 \\
        \checkmark & for ROIs & & 40.67 & 45.52 \\
        \checkmark & \checkmark & & 41.14 & 46.94 \\
        \checkmark & \checkmark & \checkmark & 41.57 & 48.63 \\
        \bottomrule
      \end{tabular*}
\end{minipage}
\hfill % Adds horizontal space between the minipages
\begin{minipage}[t]{0.49\textwidth}
      \caption{Ablation study of two types of instruction data.}
      \vspace{-1em}
      \label{tab:instruction_type}
    \centering
      \small
      \begin{tabular*}{\textwidth}{@{\extracolsep{\fill}}c@{\extracolsep{\fill}}c@{\extracolsep{\fill}}c@{\extracolsep{\fill}}c@{\extracolsep{\fill}}c@{\extracolsep{\fill}}}
        \toprule
        \multirow{2}{*}{Category} & \multirow{2}{*}{Conversation} & AffectNet & \multicolumn{2}{c}{MAFW} \\ \cmidrule(rl){4-5}
        & & Acc & UAR & WAR  \\ 
        \midrule
        & & 32.96 & 19.33 & 23.11  \\
        \checkmark & & 57.93 & 31.56 & 44.15 \\
        \checkmark & \checkmark & 59.58 & 34.75 & 44.86 \\
        \bottomrule
      \end{tabular*}
\end{minipage}
\begin{minipage}[t]{0.3\textwidth}
    \caption{Ablation study on Face Info Mining.}
    \label{tab:faceinfomini}
    \vspace{-1em}
    \centering
      \small
      \begin{tabular*}{\textwidth}{@{\extracolsep{\fill}}l@{\extracolsep{\fill}}c@{\extracolsep{\fill}}c@{\extracolsep{\fill}}}
        \toprule
        \multirow{2}{*}{Strategy} & \multicolumn{2}{c}{MAFW} \\ \cmidrule(rl){2-3}
        & UAR & WAR  \\ 
        \midrule
         Pool \& Concat& 34.51 & 44.58 \\
         Face Info Mining & 40.67 & 45.52 \\
        \bottomrule
      \end{tabular*}
\end{minipage}
\hfill
\begin{minipage}[t]{0.32\textwidth}
      \caption{Ablation study on descriptive text while inference.}
      \vspace{-1em}
      \label{tab:descriptive}
    \centering
      \small
      \begin{tabular*}{\textwidth}{@{\extracolsep{\fill}}c@{\extracolsep{\fill}}c@{\extracolsep{\fill}}c@{\extracolsep{\fill}}c@{\extracolsep{\fill}}c@{\extracolsep{\fill}}}
        \toprule
        \multirow{2}{*}{Visual} & \multirow{2}{*}{Descriptive} & \multicolumn{2}{c}{MAFW} \\ \cmidrule(rl){3-4}
        & Text & UAR & WAR  \\ 
        \midrule
        \checkmark & & 41.57 & 48.63 \\
        \checkmark & \checkmark & 49.29 & 55.42 \\
        \bottomrule
      \end{tabular*}
\end{minipage}
\hfill % Adds horizontal space between the minipages
\begin{minipage}[t]{0.35\textwidth}
      \caption{Ablation study of imbalanced dataset sampling strategy.}
      \vspace{-1em}
      \label{tab:sampler}
    \centering
      \small
      \begin{tabular*}{\textwidth}{@{\extracolsep{\fill}}c@{\extracolsep{\fill}}c@{\extracolsep{\fill}}c@{\extracolsep{\fill}}c@{\extracolsep{\fill}}c@{\extracolsep{\fill}}}
        \toprule
        \multirow{2}{*}{Resampling} & AffectNet & \multicolumn{2}{c}{MAFW} \\ \cmidrule(rl){3-4}
         & Acc & UAR & WAR  \\ 
        \midrule
        & 11.30 & 28.49 & 39.27 \\
        \checkmark & 57.93 & 31.56 & 44.15 \\
        \bottomrule
      \end{tabular*}
\end{minipage}
\vspace{-1.5em}
\end{table*}

\subsection{Implementation Details}

We initialize all the frozen weights of Emo-LLaMA with LLaMA-VID-7B~\cite{li2023llama} and FaceXFormer-B~\cite{narayan2024facexformer}, and only tune the trainable parameters during the training stage. We train our Emo-LLaMA for one epoch using a combination of category and conversation instruction data, optimized by AdamW with a learning rate of 2e-4. The rank of LoRA is set to 128. We use an imbalanced dataset sampling strategy for category instructions. The parameters for our training with LoRA adhere to the default settings established by LLaVA-1.5. For video input, we extract frames at a speed of 1 FPS. The model is trained using 8$\times$AMD MI250X GPUs. More details can be found in the \textbf{Appendix}.

\textbf{Database and Protocols.}
We perform experiments on three SFER and three DFER datasets: RAF-DB~\cite{li2017reliable}, FERPlus~\cite{barsoum2016training}, AffectNet~\cite{mollahosseini2017affectnet}, DFEW~\cite{jiang2020dfew}, FERV39K~\cite{wang2022ferv39k}, and MAFW~\cite{liu2022mafw}. We use accuracy and recall to evaluate performance for FER, following previous work. Additionally, we introduce three other multimodal emotion understanding datasets and a micro-expression recognition, performing FER in a zero-shot setting with only visual input. We hope that the experiments on these datasets could explore the generalization of MLLMs and lay the groundwork for further exploration into multimodal emotion understanding. These datasets include MELD~\cite{poria2018meld}, MER~\cite{lian2023mer}, CMMA~\cite{zhang2024cmma}, and CASME \uppercase\expandafter{\romannumeral2}~\cite{yan2014casme}.

\vspace{-0.5em}
\subsection{Experiment Results}

\textbf{Zero-shot Evaluation for Open-source MLLMs.} 
To evaluate the performance of the existing MLLMs, we compare Video-ChatGPT~\cite{maaz2023video}, Video-LLaMA~\cite{zhang2023video}, Chat-UniVi~\cite{jin2023chat}, LLaMA-VID~\cite{li2023llama}, and OneLLM~\cite{han2023onellm} in a zero-shot setting. To ensure that MLLMs, which are not trained on close-set recognition tasks, can effectively respond with the category in close-set, we use a prompt for guidance: \textit{``PLEASE ENSURE that you start your answer with `My choice is: ' FIRST and select ONLY ONE WORD from the provided list.''} This is to prevent them from generating unprocessable answers and helps align with existing metrics. The results are shown in %the 
Fig.~\ref{fig:zeroshot} and the table in the \textbf{Appendix}. We also report the performance of GPT-4V~\cite{achiam2023gpt} from ``GPT-4V with emotion''~\cite{lian2023gpt}, representing the best performance of MLLMs on FER tasks. The results indicate that existing MLLMs still lack the ability to effectively understand facial expressions and have a significant gap compared to GPT-4V. On the other hand, it demonstrates that despite being the most advanced closed-source MLLM, GPT-4V still exhibits significant performance gaps compared to specialized supervised SOTA models in handling downstream tasks such as FER.

\textbf{Comparison on FER Datasets.} 
We conduct a comparative experiment with the latest traditional methods on FER datasets, as presented in Tab.~\ref{tab:sota}. Our EMO-LLaMA achieves the SOTA-comparable or even better performance on several FER datasets in terms of accuracy and unweighted recall. The slightly poorer performance in weighted recall could be due to the use of imbalanced dataset sampling strategy for category instructions. This strategy makes the model pay more attention to the tail-end categories, which affects the weighted recall.
These results demonstrate significant potential of MLLMs in addressing the FER and emotion understanding problem. 
Despite slightly inferior performance on some datasets, we believe the possible reasons include biased emotion annotations in the dataset and factors related to training.
It is worth noting that most of supervised methods are specifically designed for FER tasks and are trained for dozens of epochs, whereas our EMO-LLaMA is easily adaptable to other tasks (e.g., emotion reasoning and multimodal emotion recognition) due to the high flexibility provided by instruction tuning and MLLMs. Additionally, our approach integrates FER with natural language, which traditional paradigms lack.

\textbf{Generalization Capability.}
We conduct cross modality zero-shot validation on image and video datasets, and zero-shot experiments on several extra datasets to verify the emotion understanding generalization capability of MLLMs on FER tasks, as shown in Tab.~\ref{tab:crossmodality_video}, Tab.~\ref{tab:crossmodality_image}, and Tab.~\ref{tab:crossmodality_other}. For cross modality validation, we perform two experiments: firstly, the model is trained on an image dataset and evaluated on a video dataset; secondly, we reverse the process. 
In the image-to-video experiment, our model outperforms GPT-4V. However, in the video-to-image setting, its performance was slightly inferior to GPT-4V. We think there are two possible reasons for this difference in generalization: i) the number of video samples is significantly smaller than that of image samples, leading to this disparity. ii) GPT-4V may have been trained on related FER data, resulting in better performance on images.
These experiments, along with those on additional datasets, demonstrate good generalization capability and suggest the potential to extend the method to multimodal emotion understanding tasks.

\vspace{-0.5em}
\subsection{Ablation Studies}

\textbf{Facial Priors.}
We explore the effectiveness of facial priors on the MAFW dataset, as shown in Tab.~\ref{tab:facial_prior}. The results demonstrate that incorporating various facial priors consistently improves overall performance. We believe this additional information includes facial structure details provided by facial embedding and landmark priors, which are lacking in the general visual encoder. We also conduct an ablation study on the Face Info Mining module, as shown in Tab.~\ref{tab:faceinfomini}. More detailed ablation study about this module can be found in the \textbf{Appendix}. When facial embedding is directly pooled and concatenated with the general embedding, the performance is similar to the baseline. This may be because pooling disrupts the fine-grained information, preventing it from effectively complementing the general embedding.

\textbf{Instruction Data Type.}
The ablation study outlined in Tab.~\ref{tab:instruction_type} provides an analysis of the impact that different instruction data types have on performance. Initially, the model, referred to as LLaMA-VID~\cite{li2023llama}, operates without the integration of the two types of instructional data and establishes a baseline for AffectNet and MAFW. This foundational performance is significantly enhanced with the inclusion of category data, which alone leads to a substantial increase in accuracy. The introduction of conversation data further amplifies this effect, underscoring the value of conversational context in enhancing the model’s predictive capabilities. This indicates that a diverse approach to instructional data significantly enhances model comprehension and performance. Therefore, to achieve more complex emotion understanding, we need to develop more sophisticated instructional datasets, such as those involving emotion reasoning.

\textbf{Descriptive Text.} 
MAFW provides short descriptive texts that include information on the environment, body movements, facial action units, and other emotional elements, which can be used for both video emotion captioning and FER. By leveraging these manually annotated descriptions, we investigate text-enhanced FER using MLLMs during inference. As shown in Tab.~\ref{tab:descriptive}, our experiment indicates that incorporating additional text significantly improves performance, an advantage that traditional paradigms lack. Our approach can easily leverage this additional information to further enhance performance, demonstrating the potential of MLLMs for improved emotion understanding.

\textbf{Imbalanced Dataset Sampling Strategy.}
We find that MLLMs also suffer from the long-tail problem when training on the classification tasks, as shown in statistic table in the \textbf{Appendix} and Tab.~\ref{tab:sampler}. To mitigate this issue, we use an imbalanced dataset sampling strategy\footnote{\url{https://github.com/ufoym/imbalanced-dataset-sampler}}. This is particularly important for AffectNet, which is a highly imbalanced dataset with a large volume of data. We notice that without this strategy, the performance after training could be worse than in a zero-shot setting or could fail to train properly.

\vspace{-0.5em}
\section{Conclusion}
In this paper, we introduce MLLMs to FER to address the limitations of current paradigms. We firstly leverage existing FER datasets to generate a large instruction dataset with Gemini, which will provide significant benefits to both the MLLM and FER communities. Furthermore, we introduce EMO-LLaMA, which effectively utilizes facial embedding, landmarks, and AGR facial priors. Additionally, we design the Face Info Mining module to extract both global and local facial information. Extensive experiments across six FER datasets demonstrate the effectiveness of EMO-LLaMA, showing its strong generalization capabilities and advantages in neural language for FER tasks. In the future, we intend to extend our method to additional emotion-related tasks, such as emotion recognition in conversations, multimodal emotion understanding, and so on. Moreover, incorporating other modalities, such as speech and audio, holds the potential for further performance improvement.

\bibliography{aaai25}

\end{document}